# Explainable AI: A Neurally-Inspired Decision Stack Framework

J.L. Olds, Schar School, George Mason University
Arlington VA 22201
jolds@gmu.edu

M.S. Khan, Schar School, George Mason University
Arlington VA 22201
mkhan63@gmu.edu

M. Nayebpour, Schar School, George Mason University
Arlington VA 22201

N. Koizumi, Schar School, George Mason University
Arlington VA 22201
nkoizumi@gmu.edu

**Abstract:**

European Law now requires AI to be explainable in the context of adverse decisions affecting European Union (EU) citizens. At the same time, it is expected that there will be increasing instances of AI failure as it operates on imperfect data. This paper puts forward a neurally-inspired framework called "decision stacks" that can provide for a way forward in research aimed at developing explainable AI. Leveraging findings from memory systems in biological brains, the decision stack framework operationalizes the definition of explainability and then proposes a test that can potentially reveal how a given AI decision came to its conclusion.

**Background:**

The recent crashes of two Boeing 737-Max commercial airliners have raised important questions about an embedded computational system (MCAS), designed to make a new model of the 737 (the Max) feel just like the older models for the human pilots (Gelles & Wichter, 2019). Among the key issues raised is that the human pilots were not informed about the existence of the system and that the system's "intelligence" was subject to a single point of failure (an angle of attack sensor) (ibid). Increasingly AI will play an important role in such systems, particularly as autonomous machines operate in remote and hostile environments such as space or deep ocean (Krichmar et al, 2019). In that context, when failures occur, it will be important to assess accurately what went wrong so that the designers can learn from failures. At the same time, when such systems make evidence-based decisions, it is important to explain why a given decision was reached. Such explainable AI has been warranted in European Union law as part of the Right to Explanation enacted in 2016, particularly in the context of adverse decisions affecting citizens.

Modern Artificial Intelligence (AI) operates on noisy and often uncertain data to make decisions on behalf of himans. When these systems work, they are of extraordinary utility allowing for, among other things, self-driving cars and autonomous robots that operate in hostile environments. Beyond utility, these systems can also engage in self-teaching modes that allows them to excel well beyond human capabilities at games like Chess and Go (Cowen, 2013; Silver, 2016 & 2017).

But, as with human intelligence, sometimes AI fails. A well-known instance of such a failure is a Tesla Model S that was involved in a fatal crash while the car was in "self-driving mode" due to inaccurate feature extraction and intelligent comprehension of a white colored truck by the AI (Shepardson, 2017).

This is not surprising: Intelligence is defined by the act of making decisions based on uncertainty and this central aspect differentiates it from non-intelligent decision systems that are based on flow-chart design (as in most computer programs) (Friedman & Zeckhauser, 2012). For human beings, during childhood development and in adulthood, such failures are required for many kinds of learning. Most machine learning (ML) AI also depends on a "training phase" whereby the artifact is instructed on a human-labeled dataset and "learns from its failures before being allowed to operate in the "wild" on non-labelled data (Sun, 2018). It is easy to understand that, in spite of training, both human and AI might mislabel a new instance of data that had never been seen before.

In the case of human intelligence only recently has neuroscience offered a clearer picture of what the cellular basis of learning and memory actually is (Liu, 2012). Neuroscience offers evidence of a concrete hierarchy with the human-body brain system at the top and neuronal synapses at the base, allowing for a framework for explaining human decisions and their concomitant failures (Fellows, 2016).

However, for AI, the explanation of why failures occur is not actually readily explainable (Gunning, 2017). This is in spite of European Union law requiring that such explanative AI be available to EU citizens to protect them from potential adverse effects of AI-based decisions such as the denial of credit (Wallace & Castro 2018, Goodman and Flaxman 2016 & 2017).

Here we propose to advance the idea of a *decision stack* as a framework for explaining AI decisions (including failures). The term is useful it reflects the idea that explanations must cross different levels of organization, in terms of complexity and abstraction (Yibo et al., 2011).

**Existence Proof: the lessons from neuroscience**

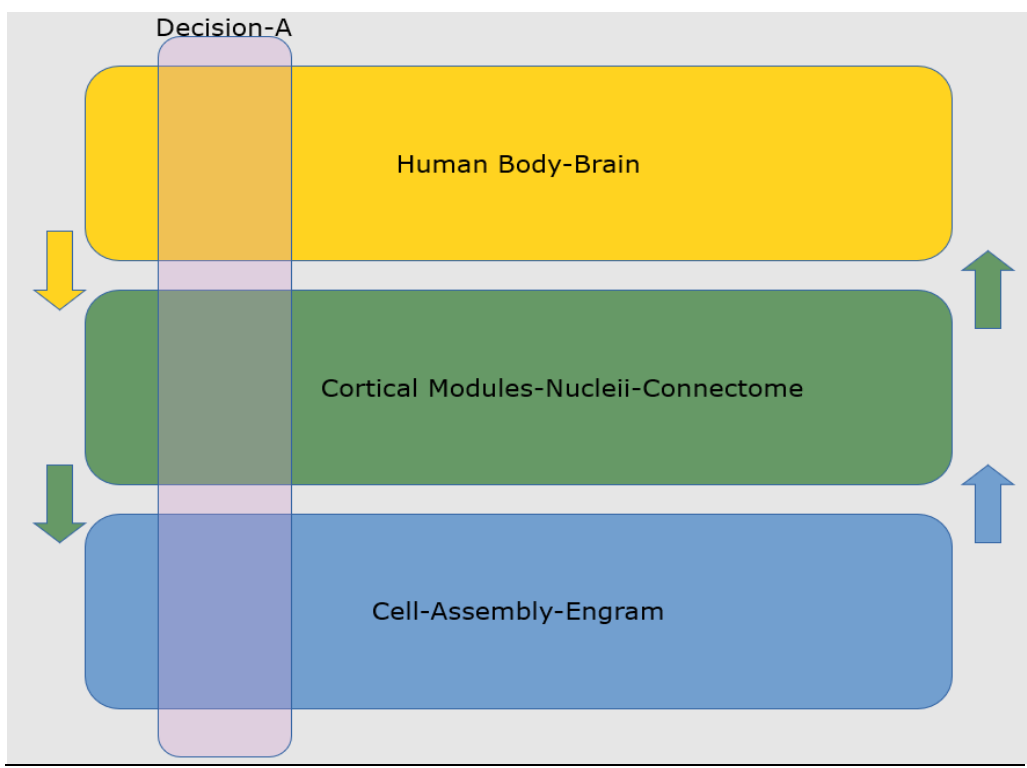

Figure 1: *The Human Brain Decision Stack. Primary control flows from the bottom to the top of the stack, albeit with top-down feedback. A set of individual neurons (c. 1000 out of 10^11) make up a cell assembly which represents a perception, concept, memory or decision. The synchronous behavior of the cell assembly results in specific activations in brain cortical modules and nuclei via short range and long-range axonal connections, known as the connectome. Those activations produce a cognitive response in the human body-brain which manifests as decisions and human behaviors.*

Since the advent of non-invasive human functional brain imaging in the last decades of the 20th century, it has become commonplace to observe the brain correlates of conscious human subjects as they make decisions at a resolution approximately 1000-fold greater in space and time than the neural code itself (Rossini, 2015; Vosskuhl, 2018; Zrenner, 2016; Pollok et al. 2015; Antal et al. 2004; Zaehle et al. 2011). These blurred images of human brains "caught in the act" of making intelligent decisions have been striking, albeit problematic from the standpoint of reproducibility. Nevertheless, clear localized neural signatures of learning and memory have been observed (Anggraini et al., 2018; Bartolotti, 2017). Impressively, AI's have been trained on such signatures and can correctly label them with appropriate nouns (e.g. cup, banana).

Blurred images of human brains in action have been connected to the lower levels of the human decision stack by numerous animal studies which record from and image individual nerve cells and even synapses (Heekeren et. al, 2004; Hsu et. al, 2005). These animal studies rely on the assumption that mechanisms of mnemonic function have been conserved across phylogeny. Operationally, neuroscience has coined the word "engram" to represent the cell assembly of neurons that participate in an individual memory (Dudai, 2004). These engrams can be clearly visualized in animal models and correlated precisely to cognitive and behavioral states in the same experimental subject.

Most importantly, novel optogenetic techniques have been developed that enable experimenter-controlled switching on and off of specific engrams to produce corresponding amnesia and subsequent rescue of memory in animal models including mouse (Ramirez, 2013). Thus, the base of the decision stack, at the level of the cell assembly, has been connected to the top of the stack, at the level of the brain-body. The intermediate components of the neurobiological decision stack correspond to various cortical modules, brain nucleii and their associated connections as illustrated in Figure 1.

The revealing of the animal decision stack allows for full explanation of animal decisions as evidenced by the optogenetic studies described above (Ramirez, 2013). By extension, such explanatory capability should be available in human subjects providing that the spatio-temporal resolution of non-invasive functional brain imaging can be extended to the level of the human neural code (engrams). While an intelligence failure can be explained by a problem in the lowest level of the stack, it might also be explained by a problem above that layer. The entire functioning stack framework is necessary for explanation of neural intelligence failure.

While the primary flow of control in human brains is from the bottom of the stack to the top, there is an additional phenomenon which adds both to the complexity and functionality of the human decision stack: feedback information and control from higher to lower levels. Such feedback is known to optimize the computational efficiency of neural computation and has been selected for over the course of evolution of brains because of energy constraints, i.e., the human brain operates on about 20 watts of electricity, the same power as a refrigerator light (Sandberg, 2016).

**Non-AI machine decision stacks**

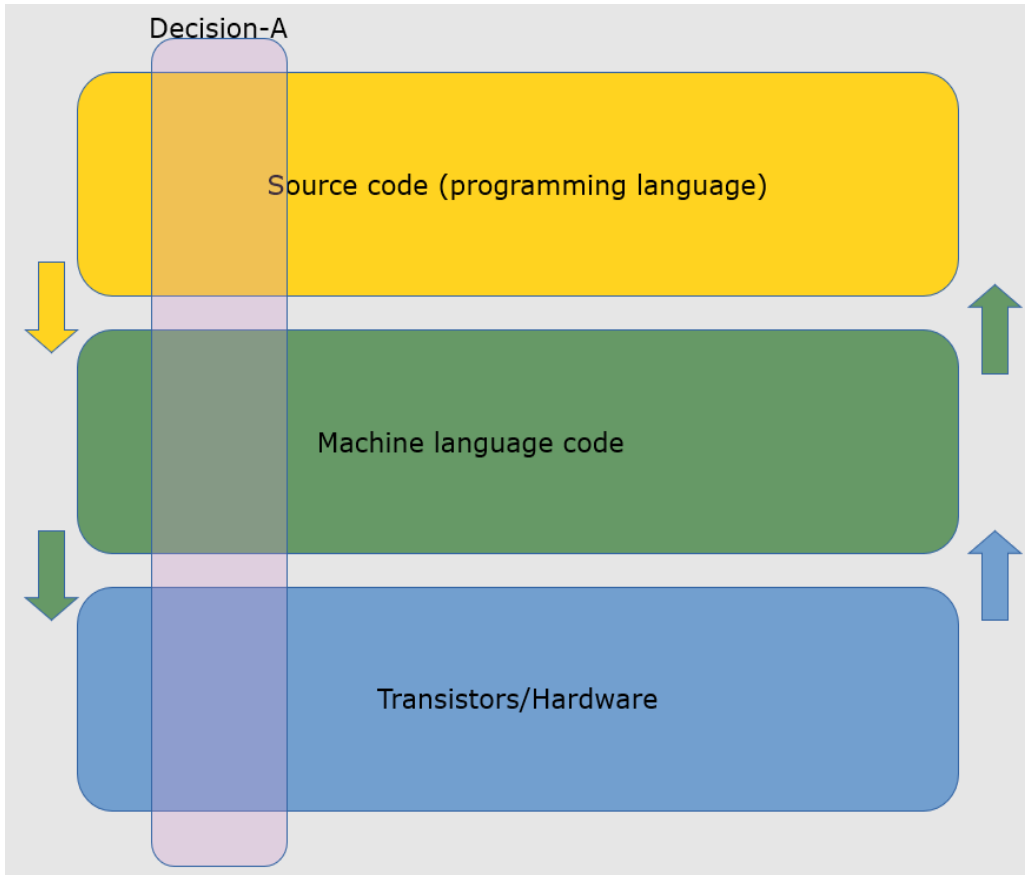

Figure 2. *The decision stack for a non-intelligent digital machine. Source code for a computer program is written in a programming language (e.g. Python). The source code describes an algorithm to operate on structured data to produce a decision. For all intents and purposes the program is deterministic. An interpreter or a compiler translates the source code into machine language which then instructs the hardware layer to turn on an off electronic components (transistors) to produce a decision which is then translated to the human operators via an interface.*

Non-AI failures occur, often manifesting on our electronic devices. When a software engineer debugs a computer program, that process reveals the explanation for failures that are embedded in the program's source code. Such de-bugging has been evolved and engineered over the years to facilitate the central roles of human beings in de-bugging. While source code has a resemblance to a written language like English, it must be translated into machine code in order to execute on a digital computer. Machine code ultimately becomes the binary sequence of 1's and 0's that drive the transistors that populate the Complementary metal–oxide–semiconductor (CMOS) circuitry of extant devices. Programing languages and debugging routines are a human-engineered tool for explaining non-AI machine failures in digital computers. However, it is the framework of the machine decision stack illustrated in Figure 2 that enables the full explanation of such failures. Source code bugs must be compiled or interpreted into machine code before they produce failure. The failure eventually manifests in the incorrect behavior of CMOS electronics.

Reading out the machine code to derive the explanation for a program failure is an alternative to current software debugging routines. However this would be very difficult as evidenced by the slow speed of debugging earliest digital computers that use machine language (Satterthwaite, 1972).

Since the machine code drives billions of transistors in modern computers, the *Gedanken* experiment of reading out each of their states to explain failure would be daunting. In contrast to the human brain, most explanation for failure comes from observing the system at the top level, not the bottom. We observe this in? a web browser that freezes on our computer desktop. The explanation is usually found in error in the source code written by a human that has propagated to the level of machine code and then to transistor hardware. Hardware malfunctions are equally capable of explaining failure modes.

For this reason, information flows in both directions in the machine decision stack found in modern devices?. For example, when the transistors inside the CPU of a modern computer become too hot, they are able to transmit that "distress signal" upwards in order to either slow down or interrupt functional execution of a program. Furthermore, many interpreters and compilers will signal coding problems upwards in the decision stack rather than blindly writing out bad machine code instructions. As a result, the arrows of information flow are bi-directional in the decision stack, as with biological brains.

**Explaining AI failures: Towards an AI decision-stack**

The challenge in providing such explanation for AI lies in the distributed nature of virtually all ML systems. A paradigmatic example can be found in artificial neural networks where computational units, "neurons", encode a decision in the pattern of their activity as influenced by the weight of their respective synaptic inputs (Yao, 1999). As with biological neurons in brains, the size and complexity of the network conceals which members are the "key players" in any given decision. As with neural systems, there are no "grandmother" cells[1] (Bowers, 2009).

An additional complexity in such networks comes from the reference architecture of many AI's where heterogenous algorithms, unstructured data storage and a separate decision engine. This is schematically represented in Figure 3.

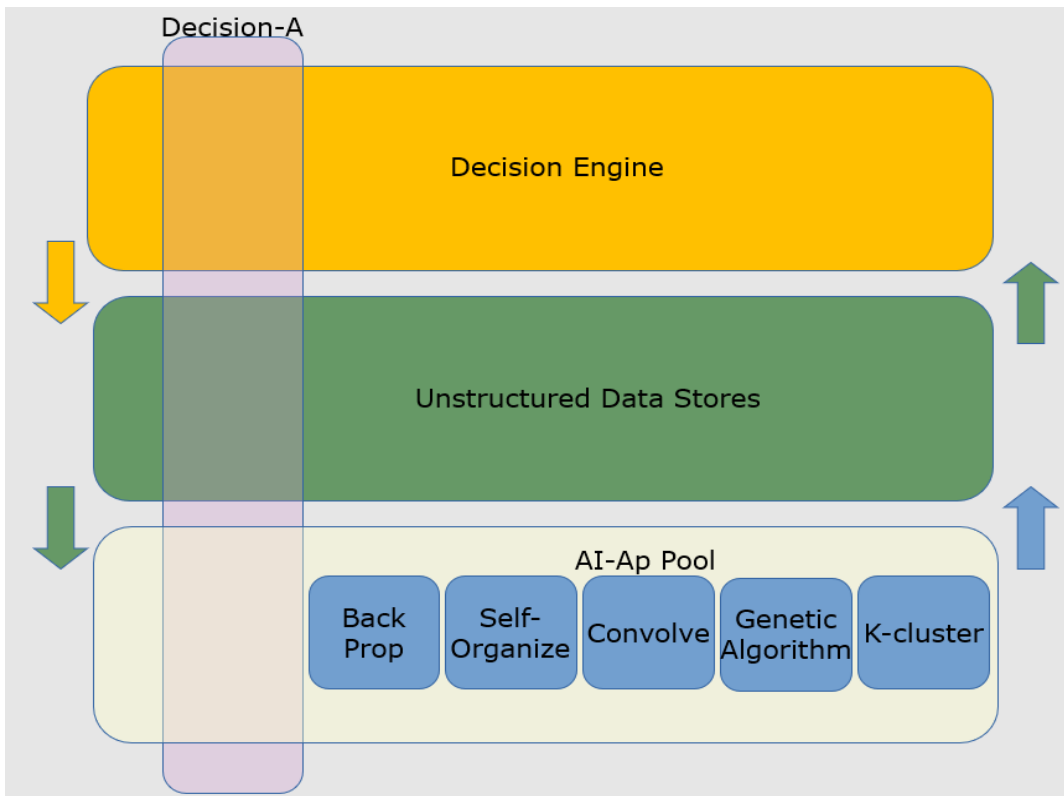

---

[1] In neuroscience, a "grandmother cell" is a theoretical construct of the single neuron that encodes the memory of your own grandmother. Evidence suggests that grandmother cells do not exist in complex biological brains.

Figure 3. *The decision stack for an AI decision machine. Pictured above is a schematized reference architecture upon which an AI decision stack (translucent pink) must operate. At the base level are homogenous or heterogenous populations of ML algorithms (AI-Ap Pool) which operate on unstructured data. The examples in the figure include back propagation artificial neural networks, self-organizing networks, convolution algorithms, genetic algorithms and K-clustering statistical methods. Data stores are implemented at the middle level often physically highly distributed because they are accessed both at the base of the reference architecture and at the top as schematized by the arrows. At the top level is a decision engine which acts as a read-out of the entire decision-stack. The decision engine acts both as read-out and also as an integral component of the decision stack that may implement some form of machine learning.*

As with biological nervous systems, the basis for a decision may be embedded in the instantiated ML networks at the base of Figure 3. It then propagate to shape the response of the top level: an *AI-decision stack*. The bottom layer of the stack in Figure 3 does not represent CMOS hardware (e.g. transistors). Rather, it represents the software-defined nodes (e.g. artificial neurons) of one or several ML algorithms. In this case, it should be possible to probe the universe of these nodes for their activity during a critical decision-making process, logging that data for future analysis in a manner analogous to the neurobiology experiments described above.

Further, and crucial to explainable AI, once such an "engram" is revealed, it should be possible, in a manner analogous to biological brains, to turn off the labeled nodes and to test whether the AI's "decision" is reversed, as propagated across the AI-decision stack. Under our operational definition of explainable AI, the results of this test constitute *the explanation*, once again taking from the field of neuroscience.

## **The Neurally-inspired Framework:**

In biological brains, the modern neuroscience approach to learning and memory is based on the notion that an *explanation* for a single biological memory is achieved by the labelling of the members of a cell assembly that is active during memory formation (i.e. an engram) and then engineering the ability to turn those, and only those cells on and off to reversibly control recall of the specific memory (Liu, 2012). Crucially, we treat mnemonic function as a generalized form of decision making and we introduce the notion of the decision stack, a biological "reference architecture". The members of the cell assembly are in the lowest layer of this reference architecture.

In the case of our AI framework, we describe an analogous decision stack reference architecture where the individual nodes/neurons are also at the lowest layer. The instrumentation of all of these nodes enables one to analogously label the relevant members to test for explanation. The test consists of re-running the decision with those nodes inactivated and revealing the dependence of the decision upon those specific nodes as they propagate their activity across the AI-decision stack.

It is important to point out that, as with biological brains, individual nodes may participate in many separate decisions and that an individually flagged node may not be crucial to explaining a single decision.

## **Conclusions:**

The future of AI's use in consequential applications will increasingly require explainability. Here we propose to use an AI-decision stack framework to operationalize AI explainability in a manner analogous to modern neuroscience: namely to instrument the complete set of ML nodes and record them all during AI decision making. The active nodes define the test set for explanation and a successful test requires demonstration of a causal relationship between the activation of the labeled nodes and the decision.

**References:**

Rossini, P. M., Burke, D., Chen, R., Cohen, L. G., Daskalakis, Z., Di Iorio, R., ... & Hallett, M. (2015). Non-invasive electrical and magnetic stimulation of the brain, spinal cord, roots and peripheral nerves: basic principles and procedures for routine clinical and research application. An updated report from an IFCN Committee. Clinical Neurophysiology, 126(6), 1071-1107.

Vosskuhl, J., Strüber, D., & Herrmann, C. S. (2018). Non-Invasive Brain Stimulation: A paradigm shift in understanding brain oscillations. Frontiers in human neuroscience, 12.

Zrenner, C., Belardinelli, P., Müller-Dahlhaus, F., & Ziemann, U. (2016). Closed-loop neuroscience and non-invasive brain stimulation: a tale of two loops. Frontiers in cellular neuroscience, 10, 92.

Pollok, B., Boysen, A. C., & Krause, V. (2015). The effect of transcranial alternating current stimulation (tACS) at alpha and beta frequency on motor learning. Behavioural brain research, 293, 234-240.

Antal, A., Nitsche, M. A., Kincses, T. Z., Kruse, W., Hoffmann, K. P., & Paulus, W. (2004). Facilitation of visuo-motor learning by transcranial direct current stimulation of the motor and extrastriate visual areas in humans. European Journal of Neuroscience, 19(10), 2888-2892.

Zaehle, T., Sandmann, P., Thorne, J. D., Jäncke, L., & Herrmann, C. S. (2011). Transcranial direct current stimulation of the prefrontal cortex modulates working memory performance: combined behavioural and electrophysiological evidence. BMC neuroscience, 12(1), 2.

Cowen, T. (2013). Average is over: Powering America beyond the age of the great stagnation. Penguin.

Silver, D., Huang, A., Maddison, C. J., Guez, A., Sifre, L., van den Driessche, G., . . . Hassabis, D. (2016) Mastering the game of Go with deep neural networks and tree search. Nature, 529(7587), 484-489).

Silver, D., et al., Mastering the game of Go without human knowledge, Nature, Vol. 550, October 19, 2017.

Shepardson, D. (2017). "Tesla Driver in Fatal 'Autopilot' Crash Got Numerous Warnings: U.S. Government - Reuters." Accessed August 21, 2019. https://uk.reuters.com/article/us-tesla-crash-idUKKBN19A2XC.

Sun, Y. (2018). More efficient machine learning could upend the AI paradigm. MIT Technology Review, February 2. https://www.technologyreview.com/s/610095/more-efficient-machine-learning-could-upend-the-ai-paradigm/

Wallace, N. & Castro,D. (2018). The Impact of the EU's New Data Protection Regulation on AI.Accessed June 23, 2018. http://www2.datainnovation.org/2018-impact-gdpr-ai.pdf.

Goodman, B. and Flaxman, S. (2017). "European Union Regulations on Algorithmic Decision-Making and a 'Right to Explanation.'" AI Magazine 38, no. 3: 50. https://doi.org/10.1609/aimag.v38i3.2741.

Goodman, B., & Flaxman, S. (2016). European Union regulations on algorithmic decision-making and a" right to explanation". arXiv preprint arXiv:1606.08813.

Yibo, C., K. Hou, H. Zhou, H. Shi, X. Liu, X. Diao, H. Ding, J. Li, and C. de Vaulx. "6LoWPAN Stacks: A Survey." In 2011 7th International Conference on Wireless Communications, Networking

and Mobile Computing, 1–4, 2011. https://doi.org/10.1109/wicom.2011.6040344.

Anggraini, D., Glasauer, S. and Klaus Wunderlich, K. "Neural Signatures of Reinforcement Learning Correlate with Strategy Adoption during Spatial Navigation." Scientific Reports 8, no. 1 (July 4, 2018): 10110. https://doi.org/10.1038/s41598-018-28241-z.

Bartolotti, J. Kailyn Bradley, K. Hernandez, E. A., and Marian, V. "Neural Signatures of Second Language Learning and Control." Neuropsychologia, Special Issue: The Neural Basis of Language Learning, 98 (April 1, 2017): 130–38. https://doi.org/10.1016/j.neuropsychologia.2016.04.007.

Ramirez, S., Liu, X, Lin, P., Suh, J. Pignatelli, M. Redondo, L. R, Ryan, J. T and Tonegawa, S. (July 26, 2013). Creating a False Memory in the Hippocampus. Science 341, no. 6144 : 387. https://doi.org/10.1126/science.1239073.

Friedman, J., Zeckhauser, Richard J., (June 7, 2012). Assessing Uncertainty in Intelligence . HKS Working Paper No. RWP12-027. Available at SSRN: https://ssrn.com/abstract=2087604

Woody, C. D. (1986). Understanding the cellular basis of memory and learning. Annual Review of Psychology, 37, 433-493. http://dx.doi.org/10.1146/annurev.ps.37.020186.002245

Fellows L.K. (2016). The Neuroscience of Human Decision-Making Through the Lens of Learning and Memory. Current Topics in Behavioral Neurosciences, vol 37. Springer, Cham

Hsu, M. (2005). Neural Systems Responding to Degrees of Uncertainty in Human Decision-Making. Science,310(5754), 1680-1683. doi:10.1126/science.1115327

Heekeren, H.R., Marrett, S., Bandettini, P.A., & Ungerleider, L.G. (2004). A general mechanism for perceptual decision-making in the human brain. Nature, 431, 859-862.

Dudai, Y. (2004). The Neurobiology of Consolidations, Or, How Stable is the Engram? Annual Review of Psychology,55(1), 51-86. doi:10.1146/annurev.psych.55.090902.142050

Satterthwaite, E. (1972). Debugging tools for high level languages. Software: Practice and Experience,2(3), 197-217. doi:10.1002/spe.4380020303

Gelles, D., Wichter, Z. (2019). Boeing 737 Max: The Latest on the Deadly Crashes and the Fallout. New York Times, July 18, 2019

Yao, X. (1999). Evolving artificial neural networks. Proceedings of the IEEE,87(9), 1423-1447. doi:10.1109/5.784219

Bowers, J. S. (2009). On the biological plausibility of grandmother cells: Implications for neural network theories in psychology and neuroscience. Psychological Review,116(1), 220-251. doi:10.1037/a0014462

Sandber, A. (2016). Energetics of The Brain and AI. Technical Report STR 2016-2 • February 2016

Gunning, D. (2017). Explainable Artificial Intelligence. DARPA/I2O, Program Update November 2017: https://www.darpa.mil/attachments/XAIProgramUpdate.pdf

Krichmar, Jeffrey L., William Severa, Muhammad S. Khan, and James L. Olds. "Making BREAD: Biomimetic Strategies for Artificial Intelligence Now and in the Future." *Frontiers in Neuroscience* 13 (2019). https://doi.org/10.3389/fnins.2019.00666.

Liu, Xu, Steve Ramirez, Petti T. Pang, Corey B. Puryear, Arvind Govindarajan, Karl Deisseroth, and Susumu Tonegawa. "Optogenetic Stimulation of a Hippocampal Engram Activates Fear Memory Recall." *Nature* 484, no. 7394 (April 2012): 381–85. https://doi.org/10.1038/nature11028.